\documentclass[a4paper, 10pt, conference]{ieeeconf}      

\IEEEoverridecommandlockouts                              

\usepackage[acronym]{glossaries}
\usepackage{amssymb}
\usepackage{siunitx}
\usepackage{isomath}
\usepackage{bm}
\usepackage[dvipsnames]{xcolor}
\usepackage[hidelinks]{hyperref} 
\usepackage{cleveref}
\usepackage{algorithm}
\usepackage[noend]{algpseudocode}
\usepackage{comment}
\usepackage{physics}
 \usepackage{cite}
\usepackage{tikz}
\usepackage{svg}
\usetikzlibrary{calc, automata, backgrounds, positioning, arrows, decorations.pathreplacing, decorations.pathreplacing, plotmarks, arrows.meta}
\usepackage{xurl}
\usepackage{pgfplots}
\pgfplotsset{compat=newest}
\usepgfplotslibrary{patchplots}
\usepackage{grffile}

\usepackage[belowskip=-10pt,aboveskip=1.5pt]{caption}
\usepackage{subcaption}
\usepackage{graphicx}

\usepackage[export]{adjustbox}
\usepackage{threeparttable}
\usepackage{graphicx}
\usepackage{etoolbox}
\makeatletter
\patchcmd{\@makecaption}
  {\scshape}
  {}
  {}
  {}
\makeatletter

   \usepackage[compact]{titlesec}
    \titlespacing{\section}{0pt}{1ex}{1ex}
    \titlespacing{\subsection}{0pt}{1.2ex}{0ex}
    \titlespacing{\subsubsection}{0pt}{0.75ex}{0ex}
\newcommand{\SZ}[1]{{\textcolor{blue}{\textbf{Notes SZ:} #1}}}
\newcommand{\NK}[1]{{\textcolor{orange}{\textbf{Notes NK:} #1}}}  
\newcommand{\BS}[1]{{\textcolor{red}{\textbf{Notes BS:} #1}}}

\newcommand{\ADDCITE}{{\bf\textcolor{orange}{[CITE?]}}}

\renewcommand{\vec}{\bm}

\crefname{figure}{Fig.}{Figs.}

\newcommand{\x}{\bm{x}} 
\newcommand{\X}{\bm{X}} 
\renewcommand{\u}{\bm{u}} 
\newcommand{\U}{\bm{U}} 
\renewcommand{\v}{\bm{v}} 
\renewcommand{\l}{\ell} 
\newcommand{\f}{f} 
\newcommand{\g}{\bm{g}} 

\newacronym{mbrl}{MBRL}{model-based reinforcement learning}
\newacronym[plural=GPs,firstplural= Gaussian processes (GPs)]{gp}{GP}{Gaussian process}
\newacronym{rc}{RC}{Radio-controlled}
\newacronym{mpc}{MPC}{Model Predictive Control}
\newacronym{mlp}{MLP}{Multilayer Perceptron}
\newacronym{rnn}{RNN}{Recurrent Neural Networks}
\newacronym{relu}{ReLU}{Rectified Linear Units}
\newacronym{gelu}{GELU}{Gaussian Error Linear Units}

\usepackage[export]{adjustbox}
\ifdefined\showColumnMargins 
    \usepackage{showframe}
    \usepackage{etoolbox}
    
    \newlength\Fcolumnseprule
    \makeatletter
    \newcommand\ShowInterColumnFrame{
    \patchcmd{\@outputdblcol}
      {{\normalcolor\vrule \@width\columnseprule}}
      {\vrule \@width\Fcolumnseprule\hfil
        {\normalcolor\vrule \@width\columnseprule}
        \hfil\vrule \@width\Fcolumnseprule
      }
      {}
      {}
    }
    
    \makeatother
    \ShowInterColumnFrame
\fi

\title{\LARGE \bf
Gradient-Based Trajectory Optimization With Learned Dynamics
}

\author{Bhavya~Sukhija, 
        Nathanael~Köhler,
        Miguel~Zamora,
        Simon~Zimmermann, \\
        Sebastian~Curi,
        Andreas~Krause,
        Stelian~Coros
\thanks{The authors are with the Department of Computer Science, ETH, Zürich, Switzerland.
        {\tt\small nate@striking.ch;  (bhavya.sukhija; miguel.zamora; simon.zimmermann; sebastian.curi, stelian.coros)@inf.ethz.ch; krausea@ethz.ch}. \vfill
        This paper has been accepted at ICRA 2023.}%
}

\begin{document}

\maketitle
\thispagestyle{empty}
\pagestyle{empty}

\begin{abstract}
Trajectory optimization methods have achieved an exceptional level of performance on real-world robots in recent years.
These methods heavily rely on accurate analytical models of the dynamics, yet some aspects of the physical world can only be captured to a limited extent. 
An alternative approach is to leverage machine learning techniques to learn a differentiable dynamics model of the system from data. In this work, we
 use trajectory optimization and model  learning for performing highly dynamic and complex tasks with robotic systems in absence of accurate analytical models of the dynamics. 
We show that a neural network can model highly nonlinear behaviors accurately for large time horizons, from data collected in only \emph{25 minutes} of interactions on two distinct robots: \emph{(i)} the Boston Dynamics Spot and an \emph{(ii)}  RC car. Furthermore, we use the gradients of the neural network to perform gradient-based trajectory optimization. In our hardware experiments, we demonstrate that our learned model can represent complex dynamics for both the Spot and \gls{rc} car,
and gives good performance in combination with trajectory optimization methods.  
\end{abstract}


\vspace{-0.5em}
\section{Introduction}\label{intro}
\vspace{-0.5em}
%
%
%
%
Robots are expected to perform complex and highly dynamic maneuvers in unknown environments~\cite{mobilerobotssurvey,Thrun2007, scaramuzza, gehring, hutter}.  Traditional trajectory-based optimal control approaches are often used for this purpose~\cite{trajectorPlanningForRobots}. Trajectory optimization methods are well established and give physically accurate trajectories which exhibit complex and dynamic behaviors \cite{skaterbots, Bern2019TrajectoryOF, Zimmermann_2019}. 

\begin{figure}[!htbp]
    \centering
    \includegraphics[width=\linewidth]{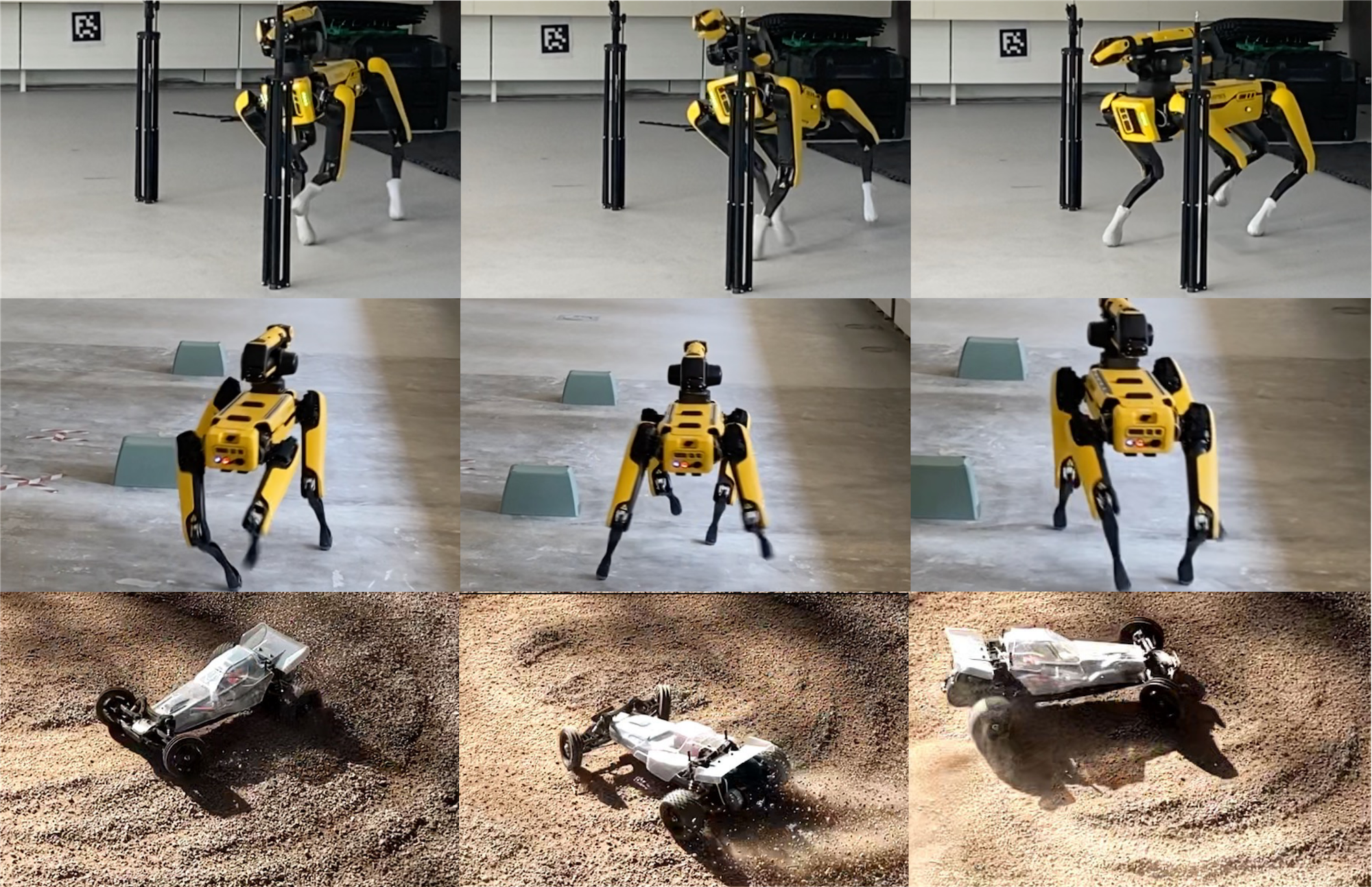}
    \caption{
    We evaluate two distinct robots Boston Dynamics Spot and a dynamic \gls{rc} car. 
    For the Spot, we consider a normal and slippery terrain. We simulate the slippery terrain by using socks at its feet. This causes the spot to slip and lose balance, as depicted on the top three images.}
    \label{fig:drift_car}
    \vspace{-1.45em}
\end{figure}

However, trajectory optimization methods require an accurate dynamics model of the system. 
Traditional modeling approaches either rely on simplified models and/or invest immense engineering effort in selecting relevant features for system identification~\cite{ASTROMSystemID, ljungSystemID, sysID, modelRobotLearning}. For highly dynamic and complex systems, it is difficult---if not impossible---to derive models following these approaches. 
For example, the Boston Dynamics Spot robot\footnote{\url{https://www.bostondynamics.com/products/spot}} has an on-board inaccessible low-level controller, and only allows control of high-level commands. This makes the Spot a complete black box and possibly non-Markovian system. Thus, deriving a dynamics model for the Spot's high-level behavior is very challenging. However, understanding its behavior is extremely essential for planning, especially when operating the robot in unknown environments (\cref{fig:drift_car}). 

\looseness=-1
The goal of this work is to leverage gradient-based trajectory optimization methods for dynamic robotic systems, such as the Spot and the \gls{rc} car, whose dynamics are unknown and challenging to model.
Towards that goal, we use data-driven methods to obtain the dynamics model.
Specifically, we record short spans of data (around 25 minutes) directly on the system and learn a model from the recorded data. We employ parametric models such as multilayer feedforward neural networks~\cite{universalapproximation} and \gls{rnn}~\cite{goodfellow2016deep} to capture the dynamics. This allows us to model the system while requiring no first principles or engineered solutions. We then leverage the learned model to perform complex and dynamic maneuvers through trajectory optimization.
This approach is evaluated on two distinct mobile robots: the Spot, and a drifting \gls{rc} car (\cref{fig:drift_car}). For the Spot, we consider two different terrains; normal/nominal terrain, and slippery terrain that we simulate by putting socks at the robot's feet (see the top half of \cref{fig:drift_car}).
In our results, we demonstrate that we can reliably learn the robot's dynamics with data collected in around $25$ minutes, and leverage the model to execute a complex trajectory using trajectory optimization. Furthermore, when operating on more slippery terrain, we show that by recording just $15$ minutes of additional data, we can adapt our model to the new surface successfully. 
Similarly, on the \gls{rc} car, we show that after collecting a small corpus of data consisting of \emph{15 minutes} of interactions, we can perform highly complex and dynamic maneuvers like drifting, which are generally difficult to model~\cite{extremecarspaper, rccarmodel}. 

\looseness=-1
Our contributions are as follows, we demonstrate that \emph{(i)} our learned model works successfully for trajectory optimization, \emph{(ii)} can be adapted for different operating conditions, 
\emph{(iii)} is capable of achieving complex drifting maneuvers, and \emph{(iv)} gives significant performance gain in lap-time for the \gls{rc} car compared to a human expert. Lastly, to the best of our knowledge, we are the first to demonstrate on the example of the Spot, a dynamic, black box, and closed-loop system, that a \gls{rnn} dynamics model can be learnt from scratch in 25 minutes and enable agile control using gradient-based trajectory optimization.
Refer to the accompanying video\footnote{\url{http://crl.ethz.ch/videos/spot_icra_compressed_final.mp4}} for more details.

\section{Related Work}
There has been a considerable amount of research in learning-based control for robotic systems~\cite{mbrlreview, PILCO, Deisenroth2015, Kamthe2018, Kabzan2019LearningBasedMP, gofetch, nagabandi2018, PETS, nagabandi2020, hucrl}.  
Most works typically use \glspl{gp}~\cite{Rasmussen06gaussianprocesses} to learn the system dynamics~\cite{chiuso_sys_id}. 
\Glspl{gp} are powerful non-parametric machine learning models that can exhibit strong theoretical guarantees, but they scale poorly for large datasets~\cite{Rasmussen06gaussianprocesses}. 
Alternatively, neural networks have been suggested as an expressive class of parametric models~\cite{SJOBERG1994359, narendraNNSysID}. Specifically, \gls{mlp}, and \gls{rnn} have shown promising results in modeling unknown nonlinear dynamical systems~\cite{rnnStudy,ogunmolu2016nonlinear, rnnstudy2}. 
Neural networks can capture complex behaviors, and therefore are often used in deep \gls{mbrl}~\cite{nagabandi2018, PETS, nagabandi2020, hucrl}. Here, generally, the methods either also learn a control policy, i.e., end-to-end control (for instance, an \gls{mlp} that outputs control signals for a given state), or use population-based search heuristics such as the cross-entropy method~\cite{CEM} for trajectory optimization. 
Nonetheless, there are notable exceptions such as~\cite{GPSNN, gradientBasedTOSim} that deploy traditional trajectory optimization solvers. In \cite{GPSNN}, local time-varying linear dynamics are learned and then integrated into an iLQG~\cite{ilq} based trajectory optimizer, which is finally used to learn a parametric policy. In contrast, we learn a global dynamics model and use a gradient-based direct shooting trajectory optimization approach for simplicity. Our approach is straightforward to implement and works successfully on two distinct and challenging mobile robots. A similar approach is used in~\cite{gradientBasedTOSim}, where trajectory optimization is performed using the out-of-the-box Adam optimizer~\cite{adam}. Specifically, their work focuses on regularizing trajectory optimization to prevent the exploitation of model inaccuracies using denoising auto-encoders. The proposed scheme is then tested in simulation. However, the objective in our work is different since we focus on the successful deployment on real hardware. Particularly, we want to demonstrate how our learned models can be successfully leveraged to optimize trajectories and deploy them on real, and dynamic robots. 
Most closely related to our approach are~\cite{ williamsModelPredictivePath2015, gofetch}.
The approach in~\cite{williamsModelPredictivePath2015} uses a neural network to learn the dynamics of a \gls{rc} car and perform dynamic maneuvers through a model predictive controller with a sampling-based optimization scheme.
However, in our work, we can achieve similar dynamic behavior using our gradient-based trajectory optimization approach, which is fast, especially in high dimensions, and known to have strong local convergence guarantees~\cite{gradientsexploding}.
In~\cite{gofetch} a model for the Spot is learned and leveraged for gradient-based trajectory optimization. However, ~\cite{gofetch} considers a parametric model with hand-picked features. For a black-box system like the Boston Dynamics Spot, hand-picking features is a time-consuming, and often unintuitive process.
We overcome this limitation by employing \glspl{rnn} instead of hand-picking features. \Glspl{rnn} are suitable for sequence modeling by design~\cite{RNN_sequence}, and have been successfully used to learn dynamic models for predicting longer time dependencies \cite{rnnStudy, rnnstudy2, ha2018recurrent, hafner2019learning, dreamer}. However, most of these works focus on simulation setups, and do not consider complex and dynamical real-world systems like the Spot. Especially on the case of the Spot, due to its low-level controller and gait cycle, the system might not be non Markovian~\cite{puterman2014markov}. However, the \gls{rnn} allows us to capture its dynamics by a learning the hidden state.
\section{Method}
\label{sec:method}
The goal of this paper is to find an optimal finite-horizon control sequence for our dynamical system.
We formulate this as a \emph{time-discretized} trajectory optimization problem:
Let $\X := (\x_1, ..., \x_n)$ and $\U := (\u_0, ..., \u_{n-1})$ be the stacked state and control input vectors for a total of $n$ trajectory steps.
Given a known initial state $\x_0$, we write the trajectory optimization problem as
\begin{align}
            &\!\!\min_{\U}  & \quad & \l(\X, \U)         
         \label{eq:trajectoryOptimizationProblem}\\ 
        & \text{s.t.} & & \x_{i+1} = \x_{i} + \f(\x_i, \u_i), \hspace{0.25cm} \forall i = 0, ..., n-1 \notag
\end{align}
with total cost $\l$ and a deterministic state transition function $\f$.
The latter models the dynamics of the physical system we want to control.
\subsection{Trajectory Optimization}
We solve the trajectory optimization problem as stated in \Cref{eq:trajectoryOptimizationProblem} through a gradient based method.
Hereby, we are interested in finding the optimal control parameters that minimize the total cost $\l(\X, \U) := \l(\X(\U), \U)$.
By following the chain rule, we can compute the gradient as
\begin{equation}
    \dv{\l}{\U} = \pdv{\l}{\X} \dv{\X}{\U} + \pdv{\l}{\U}.
\end{equation}

We then perform gradient-based optimization either using standard optimizers such as Adam~\cite{adam} or a simple line search for the step size. To avoid convergence to bad local optimas, we run the optimization with random initialisations and pick the best sequence. 

The Jacobian $\dv{\X}{\U}$ depends on the state transition function. Specifically, it is a lower diagonal matrix that can also be computed via the chain rule:
\begin{equation}
        \left[\dv{\X}{\U}\right]_{i,j} = \pdv{x_i}{u_j},
\end{equation}
\begin{equation*}
    \pdv{x_i}{u_{j}} = \begin{cases}
    \pdv{x_i}{x_{i-1}} \pdv{x_{i-1}}{u_{j}}, \quad \forall j<i-1, \notag \\
    \pdv{f}{u}\Bigl\rvert_{(x_{i-1}, u_{i-1})} \quad j=i-1,  \\
    0 \quad \hspace{5em} \text{else},
    \end{cases}
\end{equation*}
\begin{equation*}
    \pdv{x_i}{x_{j}} = \begin{cases}
    \pdv{x_i}{x_{i-1}}  \pdv{x_{i-1}}{x_{j}}, \quad \forall j< i-1, \notag \\
    1  + \pdv{f}{x}
    \Bigl\rvert_{(x_{i-1}, u_{i-1})} \quad \hspace{-2em} j=i-1, \\
    1 \quad \hspace{5.5em} j=i, \\
    0 \quad \hspace{5em} \text{else}.
    \end{cases}
\end{equation*}
\subsubsection{Control Costs}
We construct the cost function, $\l(\X, \U)$(~\Cref{eq:trajectoryOptimizationProblem}), as a sum over input penalties and state-wise immediate costs. In particular, we encourage smooth trajectories by penalizing both high magnitudes and high changes in the control inputs throughout the entire time horizon.
The corresponding cost can be written as
\begin{equation}
    \l_{\text{reg}}(\U) = w_{u} \sum_{i = 0}^{n-1} \norm{\u_{i}}^2 + w_{\text{jerk}} \sum_{i = 1}^{n-1} \norm{\u_{i} - \u_{i-1}}^2.
\end{equation}
Here, $w_{\text{jerk}}$ and $w_{u}$, are weights used to penalize jerks, and large control magnitudes respectively.
We define the state-wise immediate costs as
\begin{equation}
    \l_{\text{target}}(\X) = \sum_{i \in I} \norm{\x_i - \bar{\x}_i}^2,
    \label{eq:cost_state_targets}
\end{equation}
Here, $\bar{\x}_i$ is the predefined target state at time step $i \in I$. We select the target state according to a reference trajectory we want the system to follow.  The overall cost is a weighted sum of the two objectives, 
\begin{equation}
  \l(\X, \U) = w_{\text{target}}\l_{\text{target}}(\hat{\X}) +  w_{\text{reg}}\l_{\text{reg}}(\U).
    \label{eq:overall_cost}
\end{equation}
\subsection{Learning the Dynamics} \label{learned_car_model}
Though trajectory optimization in itself is well studied, the main challenge for us stems from the unknown dynamics $f$. To this end, we represent $f$ as a parametric model $f_{\theta}$, and learn the parameters $\theta$.
Specifically, we record a dataset of transitions $\{x_k, u_k, x_{k+1}\}$ directly on the robots and use the collected data to learn the dynamics in a supervised manner by maximizing the data likelihood. The learned model $f_{\theta}$ is then used for trajectory optimization. Then, we fix the learned model $f_{\theta}$, and leverage it to perform trajectory optimization to find the optimal control input $\U^*$:
\begin{align}
        &\!\!\min_{\U}  & \quad & \l(\hat{\X}, \U)         \label{eq:TO_cost_with_net} \\\ 
        & \text{s.t.} & & \hat{\x}_{i+1} = \hat{\x}_i + \f_{\theta}(\hat{\x}_i, \u_i), \hspace{0.25cm} \forall i = 0,  ..., n-1 \notag
    \label{eq:TO_problem_net}
\end{align}
where $\hat{\X} := (\hat{\x}_1, ..., \hat{\x}_n)$ is the concatenation of predicted $n$-step trajectory using the learned model.

\subsubsection{Learning for the Spot} In this work, we consider a three-dimensional input space for the Spot, which corresponds to the forward, sideward, and angular velocities. This allows us to move the robot's base in a $2$D plane only. Therefore, the states of the Spot we consider are the position, orientation, and velocities in the local frame, i.e., 
\begin{equation*}
    \x = [p^l_x, p^l_y, \psi, v^l_x, v^l_y, \Dot{\psi}]^T.
\end{equation*}
Furthermore, when considering mobile robots operating on homogeneous terrains, we can assume that the dynamics are invariant with respect to the robot's global position. Therefore, we do not consider the global positions as input for $f$ when learning the dynamics. This reduces the input space for $f$ and potentially helps in faster generalization. 

We control the Spot at a \SI{20}{\hertz} frequency and fix its gait to \emph{trot}. 
For the state measurements, we use the on-board Spot state estimator.
 From initial experiments, we notice that the leg joint configuration and gait cycle of the Spot influences its behavior. 
 This influence is not captured by description of our state space. 
 To this end, we learn a hidden state by using an \gls{rnn}, specifically a gated \gls{rnn} (GRU)~\cite{GRU}, with the hope that the hidden state can represent the true dynamics of the robot better. Moreover, we deliberately choose a simple state-space representation of the Spot and leverage the learned hidden state to compensate for other unaccounted influences.  

\subsubsection{Learning for the RC car}  We use an \gls{rc} car with a high torque motor, which allows 
us to perform dynamic maneuvers that involve loss of traction and drifting. The state of the car consists of three degrees of freedom, two for its position, and one for its orientation. This corresponds to the same state space as the Spot. The inputs for the car are the forward velocity and steering angle. We use the Optitrack for robotics motion capture system\footnote{\url{https://optitrack.com/applications/robotics/}}.
We use a feed-forward neural network to capture the dynamics.

\subsubsection{Regularization and Continuous Activation Functions} \label{subsubsec:activationfunction} 
Since we use our learned model for gradient-based trajectory optimization, we prefer smooth derivatives. Smooth derivatives not only ease the trajectory optimization itself but also result in smoother action sequences.
To this end, 
we choose continuous activation functions, such as \gls{gelu}~\cite{GELU} for our neural networks and apply a L2 regularization to avoid overfitting. 
Our approach is summarized in \Cref{alg:main}.
\vspace{-1em}
\begin{algorithm}
\caption{Learned Model for Trajectory Optimization}
\label{alg:main}
\begin{algorithmic}
\Require Data: $\mathcal{D}$, Initial Input Sequence $U_{0}$ 
\State Train model from dataset $\mathcal{D}$: $\theta^{*} \leftarrow \min_{\theta} L(\mathcal{D}|f_{\theta})$.
\State $k \leftarrow 0$
\While{Not Converged and $k<N_{\max}$} 
    \State $U_{k+1} \leftarrow U_k - \eta_{k} \dv{\hat{\l}_{\theta}}{\U}$
    \State $k \leftarrow k + 1$
\EndWhile
\Return{$U_{k}$}
\end{algorithmic}
\end{algorithm}
\vspace{-1.5em}

\section{Results}
\label{sec:results}
\begin{figure*}[!htbp]
\centering
 \includegraphics[width=\linewidth]{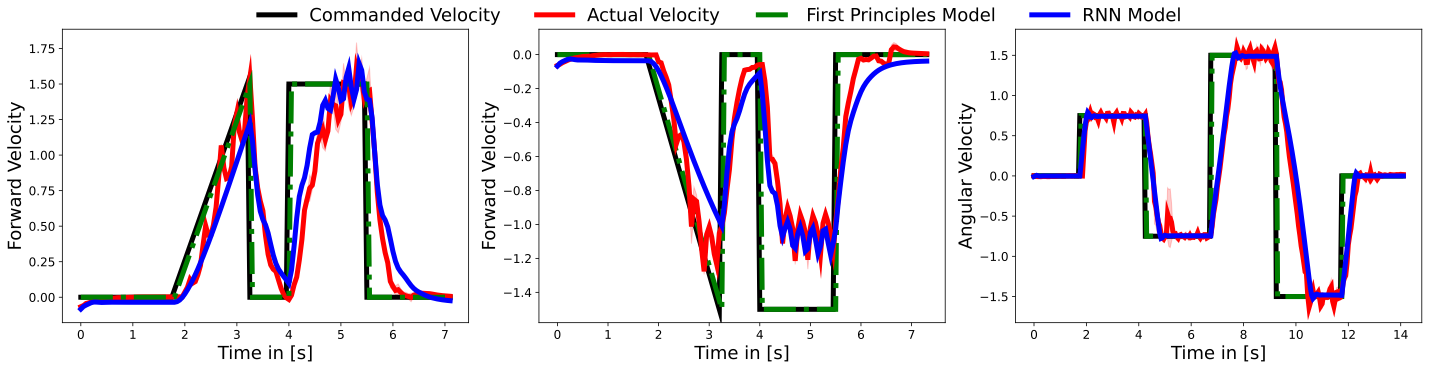}
    \caption{Spot's behavior on normal terrain (red) vs predictions using the first principle model (green), and the learned model (blue) over $3$ independent runs for forward, backward, and turning motion commands (black-dashed line). }
    \label{fig:comaprison_spot}
\end{figure*}
This section presents experimental results on hardware achieved through trajectory optimization using learned models. The aim of our experiments is to demonstrate that our learned models can \emph{$(i)$}  capture nonlinear dynamics of the robots well, and \emph{$(ii)$} be successfully used for trajectory optimization. Therefore, we learn models for two mobile robots, the Spot and the \gls{rc} car. To demonstrate the strength of our models, we perform open-loop trajectory optimization. Lastly, we demonstrate how our learned model can be successfully integrated into closed-loop trajectory optimization on the example of the \gls{rc} car on a race track.
 A summary table of our open-loop trajectory optimization results is presented in \Cref{tab:offlineTrajComparison}. 
We also provide a video (see~\Cref{intro}) of our dynamic motions on hardware.
\subsection{Boston Dynamics Spot Experiment} \label{spot}
\begin{figure}[!htbp]
    \centering
    \begin{subfigure}[b]{0.49\linewidth}
       \includegraphics[width=1.1\linewidth]{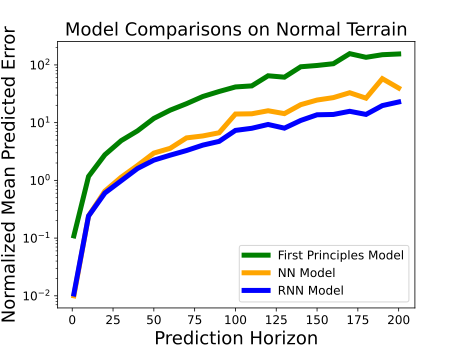}
    \end{subfigure}
    \begin{subfigure}[b]{0.49\linewidth}
       \includegraphics[width=1.1\linewidth]{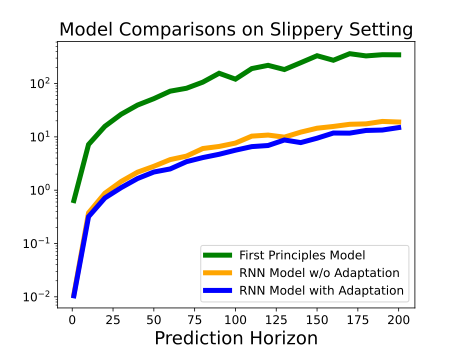}
    \end{subfigure}
 \caption{\looseness=-1
 Left: Normalized mean prediction error accumulation, in log-scale, over multiple horizon lengths for simple model, neural network model, and RNN model for the normal terrain. Right: Normalized mean prediction error accumulation, in log-scale,  over multiple horizon lengths for the simple, unadapted, and adapted model for slippery terrain.}
    \label{fig:RNN_comparison}
\end{figure}

Due to the black-box nature of the Spot, it is difficult to derive a model from the first principles. A simple model one may consider is $v^l_{x,k+1}= u^{\textit{Forward}}_k$, $v^l_{y,k+1} = u^{\textit{Sideward}}_k$, and $\Dot{\psi}_{k+1}= u^{\textit{Turning}}_k$, i.e., desired/commanded velocities are equal to the actual velocities of the robot. From this, the positions can be determined using Euler forward integration. This model is compared to our learned model on test trajectories with forward and turning motions (\cref{fig:comaprison_spot}). From  \cref{fig:comaprison_spot}, we can deduce that our learned model is considerably better than the simple first principles model. For instance, it is noticeable from the figure that given the low-level controller, the robot cannot walk backward as fast as commanded. Particularly, it can walk forward faster than backward. While the simple model cannot capture this behavior, our learned model can. This highlights the importance of learning a dynamics model for the Spot and also showcases the limitations of our first principle model. On the left in \cref{fig:RNN_comparison}, we compare the test-error accumulation over open-loop predictions for varying horizons between \emph{(i)} simple model, \emph{(ii)} neural network model, and \emph(iii) RNN (GRU) model. The errors of the simple model increase drastically with the horizon length. Nonetheless, the neural network model and the GRU model show better performance, with the GRU giving better results. 

\begin{figure}[!htbp]
    \centering
    \includegraphics[width=\linewidth]{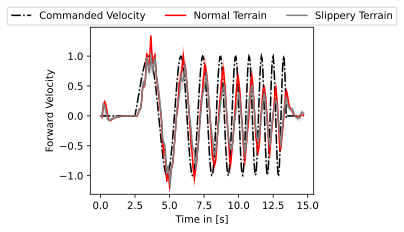}
    \caption{Tracking performance of the Spot on normal and slippery terrain (Spot with socks). }
    \label{fig:nominal_vs_slippery}
\end{figure}

To further demonstrate the benefits of learning for the robot, we simulate a slippery terrain by putting socks at the feet of the Spot. The socks cause the Spot to slide and therefore slip (\cref{fig:drift_car}). As depicted in~\cref{fig:nominal_vs_slippery}, this leads to a slight difference in Spot's tracking performance. We capture this change in dynamics, by recording another dataset for the slippery case for 15 minutes and adapting the learned model from before by retraining on the new dataset.  On the right-hand side of \cref{fig:RNN_comparison}, we compare the adapted model with the unadapted one for test data recorded on the slippery surface. From the figure, we can conclude that the adapted model performs slightly better than the unadapted one as the prediction horizon increases.

\subsubsection{Trajectory Optimization}
\begin{figure}[!htbp]
\centering
\includegraphics[width=\linewidth]{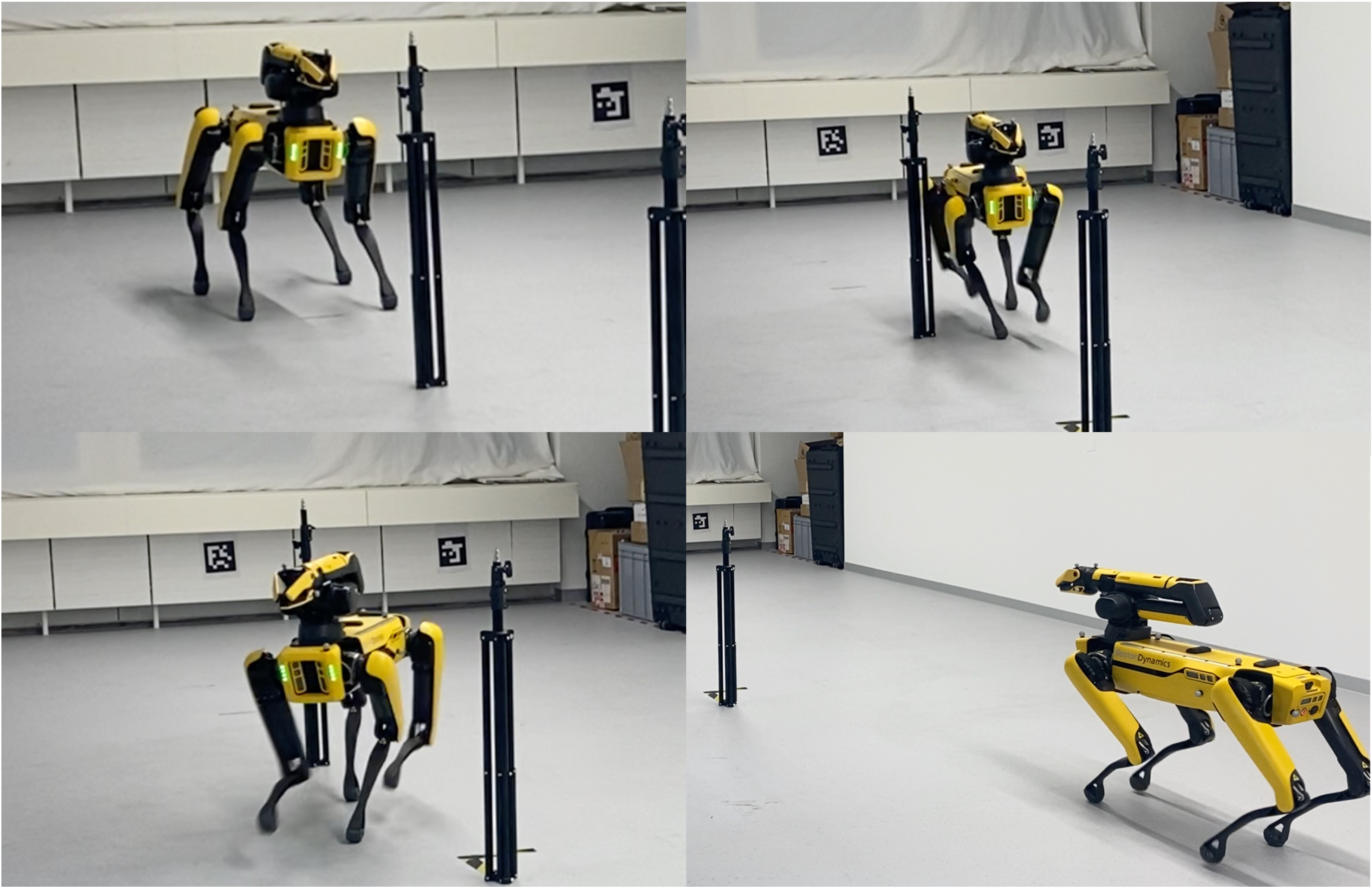}
    \caption{Spot open-loop zig-zag trajectory obtained through trajectory optimization with the learned model.}
    \label{fig:spot_traj}
        \vspace{-1em}
\end{figure}

\begin{figure}[!thbp]
\begin{subfigure}{0.49\linewidth}
     \centering
    \includegraphics[width=\linewidth]{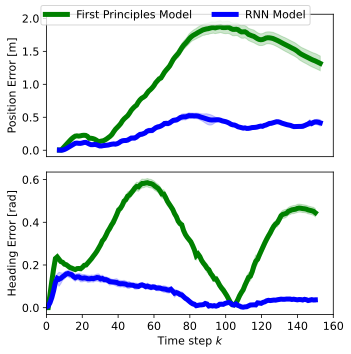}
\end{subfigure}
  \begin{subfigure}{0.49\linewidth}
     \centering
    \includegraphics[width=\linewidth]{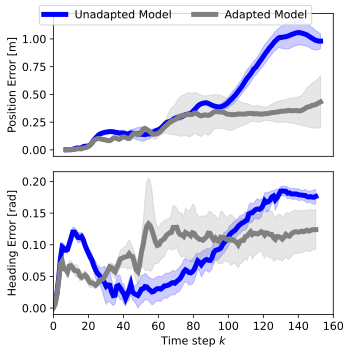}
\end{subfigure}
\caption{Comparisons for Spot experiments. Left: Prediction error using the first principles model (green) and the learned model (blue).
Right: Prediction error of the unadapted model (blue) and the  model adapted for the slippery floor (grey).
For both cases, we average over four independent trajectories and also depict the standard deviation.}
\label{fig:spot_comparisons}
  \vspace{-0.5em}
\end{figure}

We leverage our learned model to perform trajectory optimization. In order to quantify the prediction strength of our model, we execute an open-loop rollout and measure the deviation between the expected and observed trajectory. The Spot has a very good low-level controller, however because we consider an open-loop input sequence, its motion deviates considerably from the desired trajectory.
For our experiments, we provide a sinusoidal motion as a reference to execute a dynamic zig-zag drill with the Spot (\cref{fig:spot_traj}). The horizon for this trajectory is $150$. Therefore, for the open-loop execution, an accurate model is required to avoid the accumulation of errors over the horizon length. We execute the same trajectory four times. Furthermore, we perform trajectory optimization using the first principles model and compare its performance to our learned one. In~\cref{fig:spot_comparisons}, we compare the performance of the two trajectories. Specifically, we depict the error between the predicted and real trajectory. 
Our results show that the learned model performs considerably better than the first principle one, i.e., has considerably (around a factor of five) lower errors.  

\looseness=-1
We perform the same experiment for the slippery setting. Here, we compare the trajectory of our adapted model to the unadapted one, i.e., the model solely trained on nominal/normal terrain. \Cref{fig:spot_comparisons} compares the performance of the two trajectories. Our results show that overall we achieve smaller errors when using our adapted model. Furthermore, we notice that the standard deviation in our executed trajectories on the slippery terrain is higher. We believe this is due to the slipping of the robot. 
\begin{figure*}[th]
    \centering
    \includegraphics[width=1\linewidth]{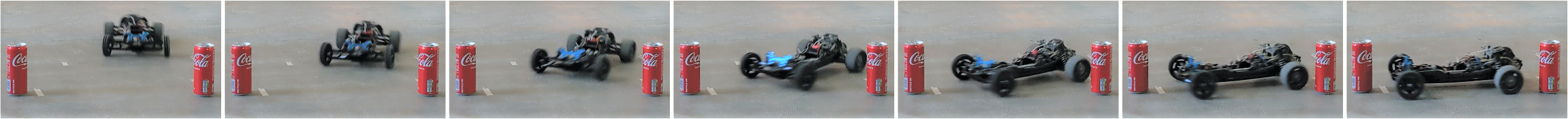}
 \caption{Executed trajectory for the parallel parking scenario.}
    \label{fig:physical_parallel_parking}
\end{figure*}
\begin{figure*}[!htbp]
\centering
\includegraphics[width=\linewidth]{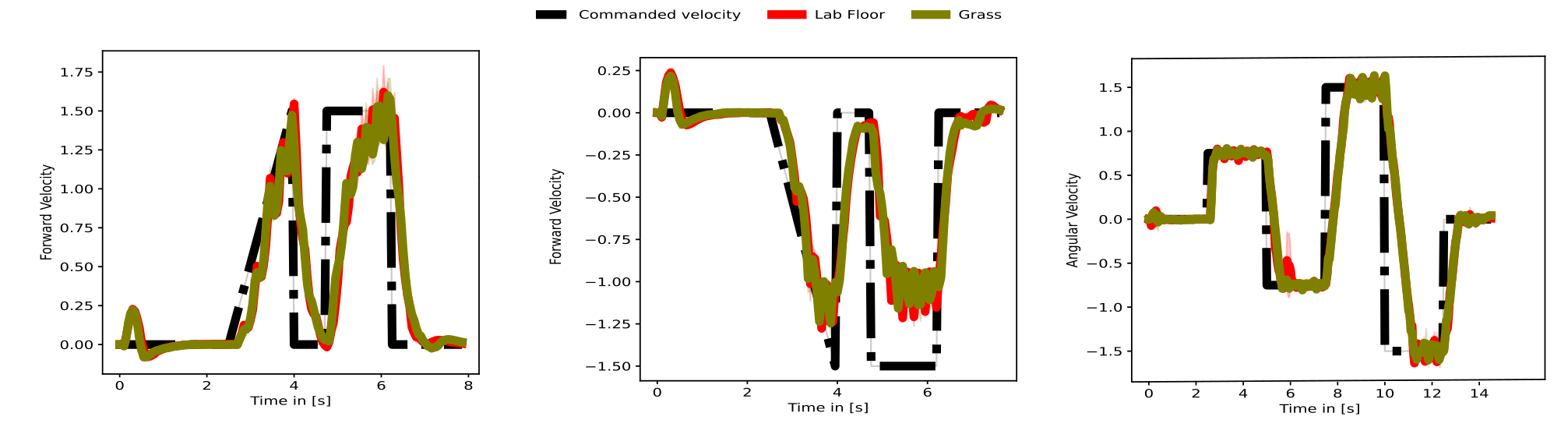}
\caption{Comparison of Spot's behavior in the normal/nominal floor (in the lab), and dry grass terrain.}
\label{nominal_vs_grass}
\end{figure*}

\subsection{RC Car} 
For the \gls{rc} car, we execute multiple open-loop rollouts. Furthermore, we evaluate the performance of our model in closed-loop using model predictive control~\cite{mpc}.
\subsubsection{Trajectory Optimization} \label{offlineTO}
\begin{figure}[ht]
\begin{subfigure}[c]{0.49\linewidth}
   \centering
   \includegraphics[width=0.4\linewidth, angle=180]{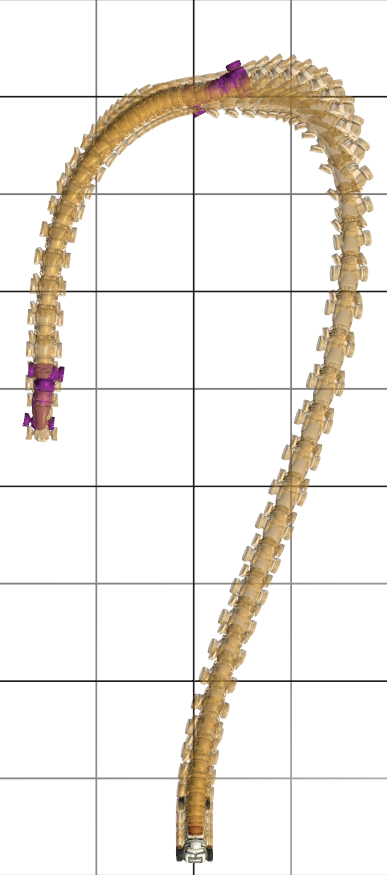}
\end{subfigure}
\begin{subfigure}[c]{0.49\linewidth}
   \centering
   \includegraphics[width=0.64\linewidth]{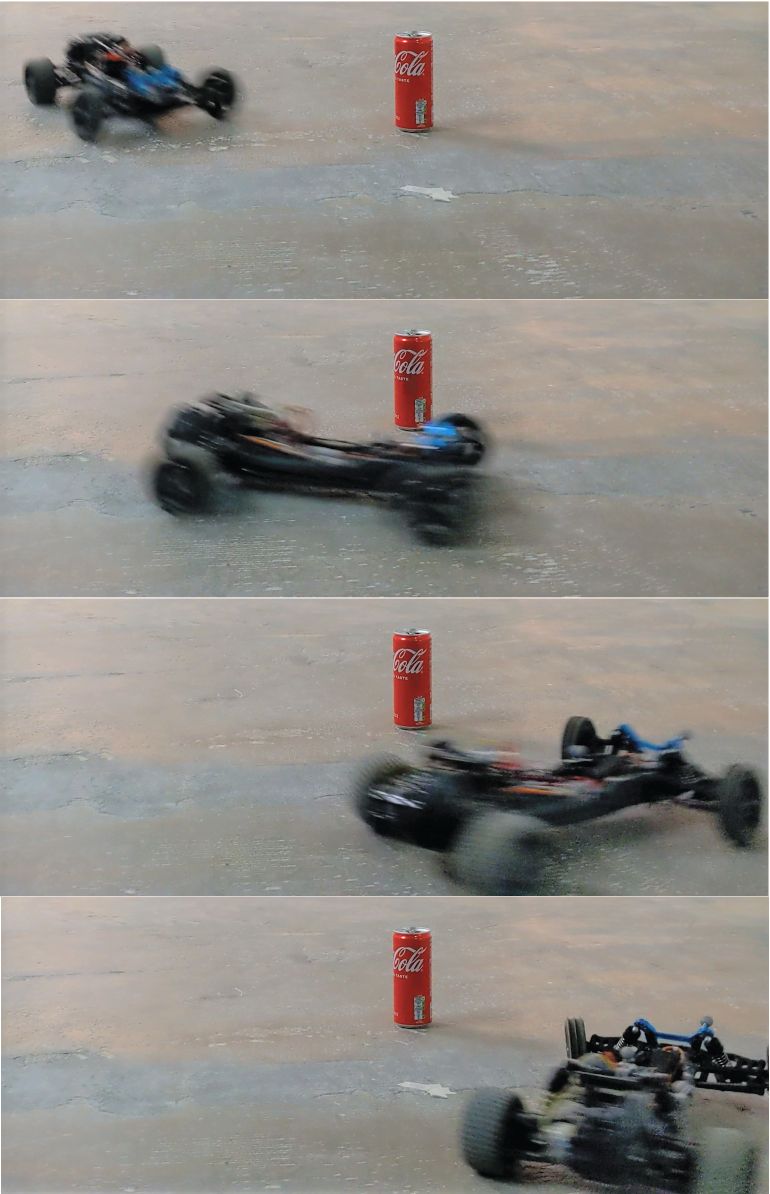}
\end{subfigure}
      \caption{The trajectory for the drifting turn scenario.}
    \label{fig:offline_scenarios}
\end{figure}
We perform trajectory optimization for three different scenarios \emph{(i)} parallel parking (\cref{fig:physical_parallel_parking}), \emph{(ii)} dynamic reverse, and \emph{(iii)} drifting turn (\cref{fig:offline_scenarios}). 
All three scenarios include dynamic drifting maneuvers.
For each scenario, we repeat the same experiment 20 times for slightly different starting positions. 
The results are summarized in \cref{tab:offlineTrajComparison}.

\begin{table}[h]
\centering
\vspace{1.75em}
\caption{The error between the planned and achieved position for open-loop trajectory optimization.}
\label{tab:offlineTrajComparison}
\begin{adjustbox}{max width=\linewidth}
\begin{threeparttable}
\begin{tabular}{lcllll}
                                                                                  &
                                                   \multicolumn{1}{l}{\begin{tabular}[c]{@{}l@{}}Spot \\ zig-zag\\
                  normal terrain      \end{tabular}} &                     
                  \multicolumn{1}{l}{\begin{tabular}[c]{@{}l@{}}Spot \\ zig-zag \\
                  slippery terrain
                  \end{tabular}} &  
                  \multicolumn{1}{l}{\begin{tabular}[c]{@{}l@{}}Parallel \\ Parking\end{tabular}} & \multicolumn{1}{l}{\begin{tabular}[c]{@{}l@{}}Dynamic \\ Reverse\end{tabular}} & \multicolumn{1}{l}{\begin{tabular}[c]{@{}l@{}}Drifting \\ Turn\end{tabular}} \\\hline
                                                                  
\multicolumn{1}{l}{\begin{tabular}[c]{@{}l@{}}Trajectory \\ Length\end{tabular}}                                                                &   150 & 150 & 40               &      40           &     60 \\
\multicolumn{1}{l}{\begin{tabular}[c]{@{}l@{}}Mean L2 \\ Position Error {[}m{]}\end{tabular}}         &   0.31 & 0.24 & 0.38               &     0.21            &     0.37          \\
\multicolumn{1}{l}{\begin{tabular}[c]{@{}l@{}}STDev L2 \\ Position Error {[}m{]}\end{tabular}}        &  0.03 &  0.09& 0.10               &       0.074            &       0.21         \\
\multicolumn{1}{l}{\begin{tabular}[c]{@{}l@{}}Mean Absolute \\ Heading Error {[}rad{]}\end{tabular}}  &        0.07 & 0.01 &  0.13        &        0.50         &        0.10       \\
\multicolumn{1}{l}{\begin{tabular}[c]{@{}l@{}}STDev Absolute \\ Heading Error {[}rad{]}\end{tabular}} &         0.01 & 0.03 &  0.08       &         0.10        &         0.10     
\end{tabular}
\end{threeparttable}\end{adjustbox}
    \vspace{-1em}
\end{table}

\begin{figure} [!htbp]
    \begin{subfigure}{0.49\textwidth}
       \centering
       \includegraphics[width=0.8\linewidth]{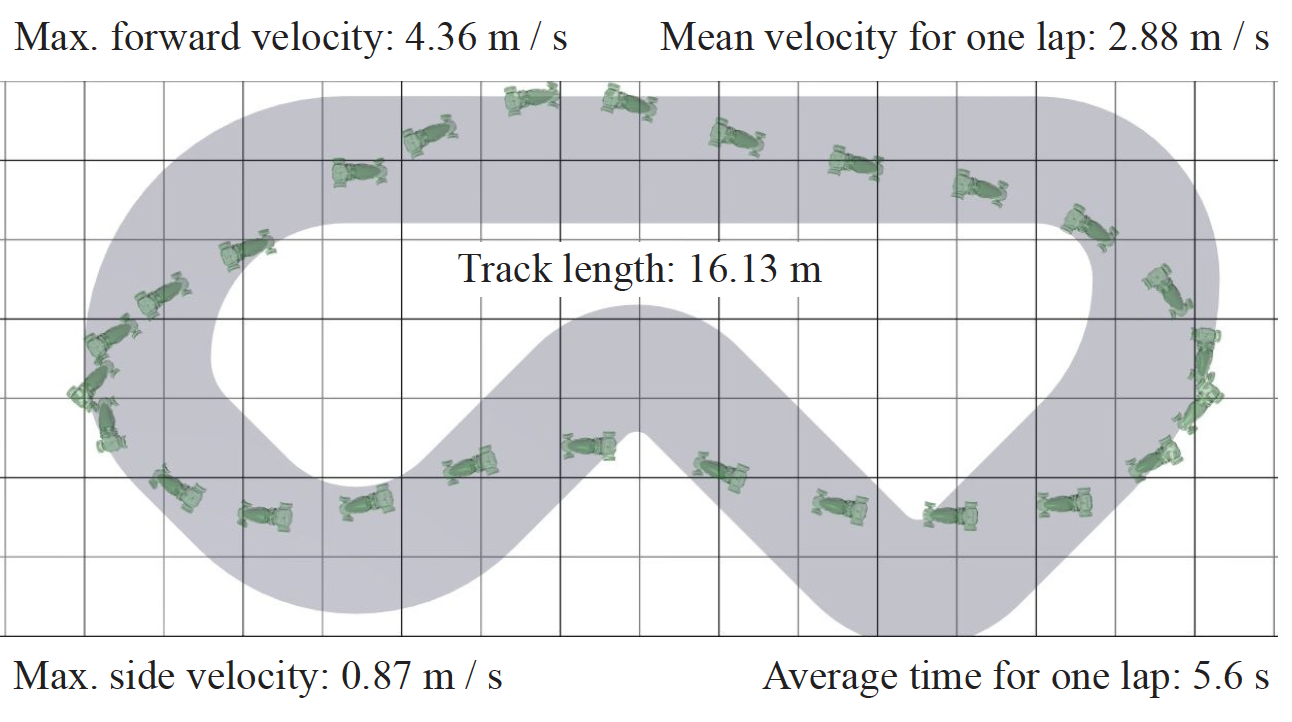}
    \end{subfigure}
    \begin{subfigure}{0.49\textwidth}
       \centering
       \includegraphics[width=0.8\linewidth]{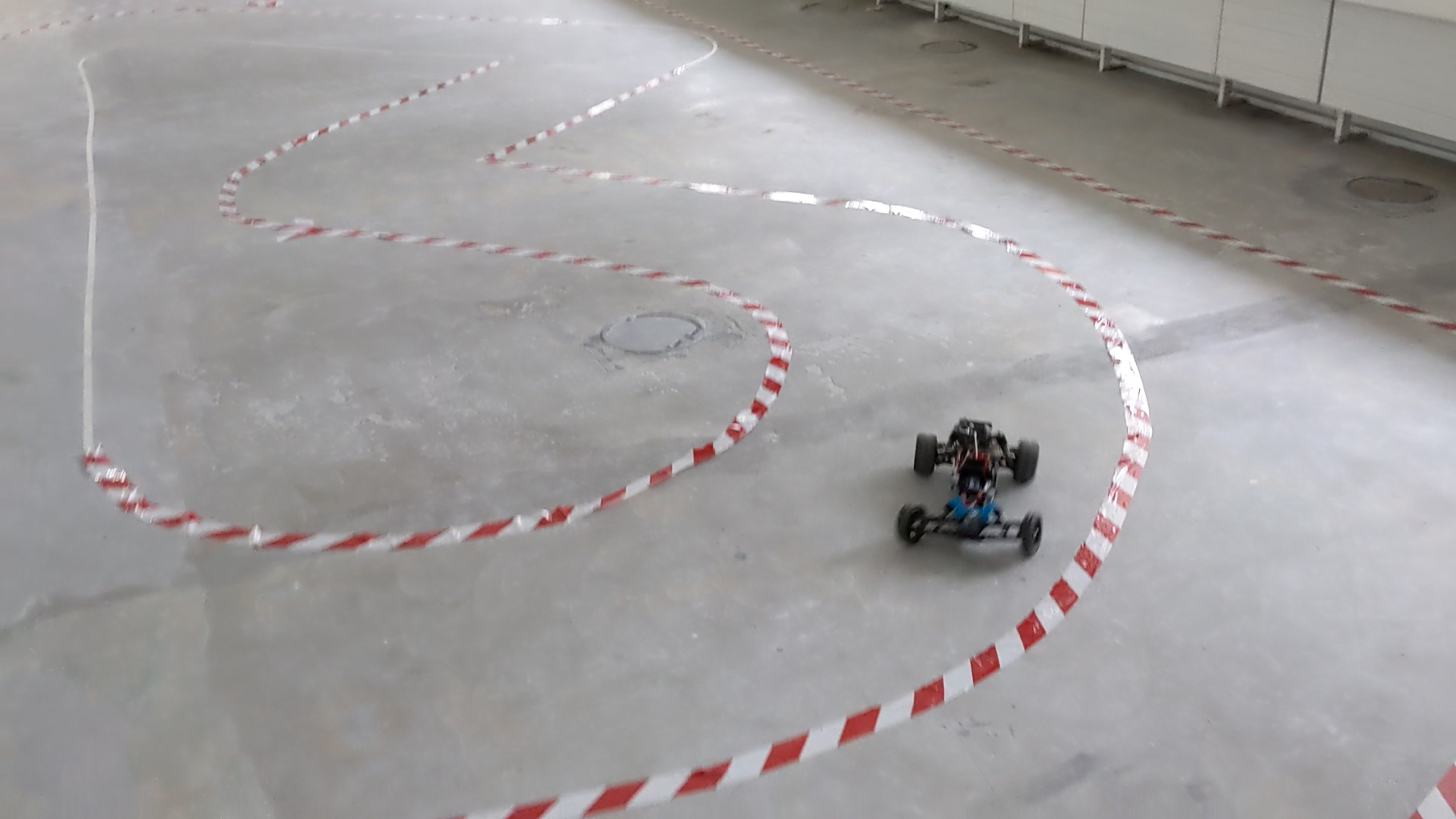}
    \end{subfigure}
    \caption{The physical \gls{rc} car (below) with feedback controller.}
    \label{fig:track_recorded}
    \vspace{-1.1em}
\end{figure}

\subsubsection{Closed-Loop Trajectory Optimization} \label{closedTO}
We perform experiments in an online trajectory optimization setting by driving on a predesigned race track (\cref{fig:track_recorded}). Closed-loop trajectory optimization is the more conventional approach to controlling robots. Thus, in this experiment, we demonstrate how it can be successfully achieved with our learned model. 
To this end, we apply trajectory optimization in a receding horizon fashion. 
The overall objective (introduced as costs $\l(\X, \U)$) is to advance along the race track with a predefined high velocity for the entire time horizon of the trajectory. Additionally, we add a cost for track excursions.
Hereby, we choose a horizon of $n = 20$ and control the robot at a frequency of $20$ Hz.
We leverage parallelization to estimate the derivatives $\dv{\hat{\X}}{\U}$ with finite differences for this particular experiment.
Additional performance metrics of the RC car on the racetrack are given in \cref{fig:track_recorded}.
As shown in the video, the car is able to race through the track with high velocity while performing dynamic maneuvers.
\subsection{Discussion}
\begin{figure} [!htbp]
    \centering
    \includegraphics[width=0.75\linewidth]{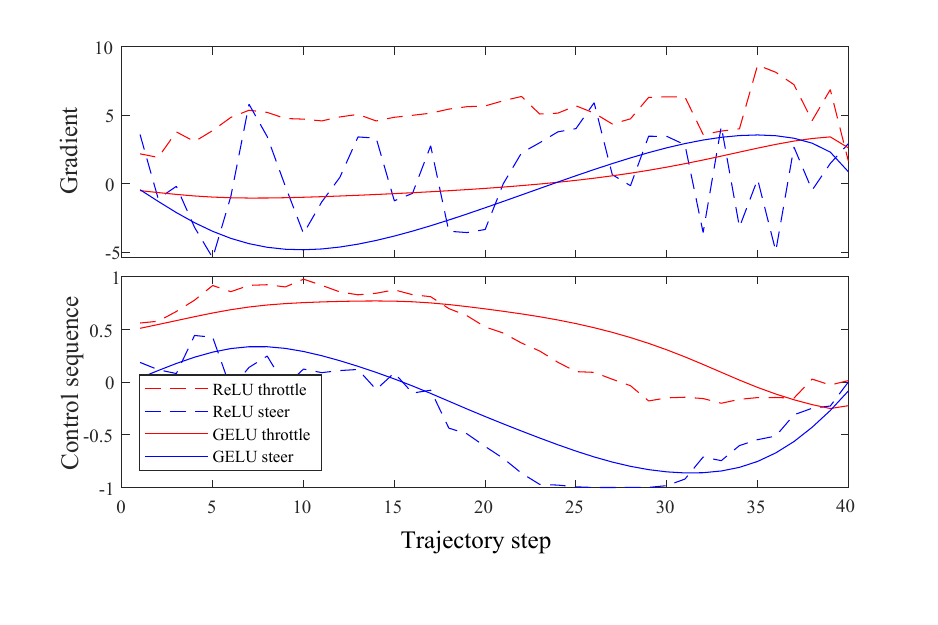}
    \caption{Comparison of gradients $\dv{\l}{\u}$ and control sequences $\u$ for neural network architectures with \gls{relu} and \gls{gelu} activations. \gls{relu} activations results in noisier (i.e. more fluctuating) gradients and input sequences than the network with \gls{gelu} activations. }
    \label{fig:smoothness}
    \vspace{-2em}
\end{figure}

\looseness=-1
In our open-loop trajectory optimization experiments, we 
notice that model inaccuracies accumulate over the trajectory horizon (\Cref{tab:offlineTrajComparison}). Even though these inaccuracies are small, they can still have an impact on the robot's performance.  Nonetheless, we can compensate for these inaccuracies using feedback control, as  we demonstrate this on the example of the \gls{rc} car~\Cref{closedTO}. 
Furthermore, during our model selection process (\Cref{subsubsec:activationfunction}), we hypothesized that paying close attention to regularization and selection of activation functions would help in obtaining smoother action sequences. Clearly, smoother action sequences are preferred when deploying directly on real-world hardware. 
We validate this hypothesis on the \gls{rc} car example.
As depicted in \cref{fig:smoothness}, the control sequence resulting from the network with \gls{gelu} activations is considerably smoother than the one obtained using \gls{relu} activations.
We trace this back to the gradient used for trajectory optimization, which is much noisier for \gls{relu} as well. Lastly, for the Spot experiments, we simulate a drastic shift in the robot's operating condition in form of the slippery terrain. 
However, given the robot's state-of-the-art low-level controller, we expect it to perform reasonably well in settings where the shift in the operating conditions is not drastic. To this end, we test the Spot on dry grass and notice that at least for forward, backward, and turning motions, the tracking performance on the lab floor and dry grass is equally good, see~\cref{nominal_vs_grass}.
Thus, we believe in such settings, we can still leverage the model learned in the lab environment.

\section{Conclusion and Future Work}
\label{sec:discussion}
\looseness=-1
The goal of this work is to leverage traditional trajectory optimization approaches for systems with unknown dynamics. We demonstrate that this can be achieved through machine learning on two distinct and challenging robots. Specifically, our results show that we can capture the dynamics of the robots, adapt our learned model to new operating conditions, and perform dynamic maneuvers using trajectory optimization. 

\looseness=-1
This work opens various avenues for further research. For instance, the exploitation of model inaccuracies by policy optimizers has been investigated in the literature~\cite{gu2016continuous, mbpo, PETS}. Suggested strategies are the use of probabilistic ensembles~\cite{rajeswaran2016epopt, kurutach2018model, pmlr-v87-clavera18a, PETS, hucrl}, shorter task horizons~\cite{mbpo} and denoising autoencoders~\cite{gradientBasedTOSim}. Since in our work we observed an accumulation of model inaccuracies (\Cref{tab:offlineTrajComparison}), in the future these approaches can be integrated to study their influence on performance. Additionally, 
in this work, the data used to learn the model was recorded offline. 
However, methods such as~\cite{hucrl} automate the data acquisition by exploring the system dynamics in an episodic online learning setting. Future work may consider leveraging these advances. 
Furthermore, in this work, we were interested in finding smooth rather than accurate gradients. 
We think studying the influence of model selection on learning accurate dynamics and gradients, as well as leveraging structured learning techniques for capturing robot dynamics~\cite{lagrangianNNs}, is an exciting direction for future work.

\bibliographystyle{IEEEtran}
\bibliography{root.bib}

\end{document}